\documentclass[11pt]{article}
\usepackage{authblk}

\newlength{\defbaselineskip}
\usepackage{multirow}
\usepackage{booktabs}
\usepackage{url}
\usepackage{graphicx}
\usepackage{amsmath}
\usepackage{amssymb}
\usepackage[caption=false,font=footnotesize]{subfig}

\usepackage{algorithm}
\usepackage{algpseudocode}

\begin{document}

\title{
Aerial Image Classification in Scarce and Unconstrained Environments via Conformal Prediction
}

\author{Farhad Pourkamali-Anaraki\\Department of Mathematical and Statistical Sciences, University of Colorado Denver, CO, USA\\ Email: farhad.pourkamali@ucdenver.edu}

\date{\today}

\maketitle

\begin{abstract}
This paper presents a comprehensive empirical analysis of conformal prediction methods on a challenging aerial image dataset featuring diverse events in unconstrained environments. Conformal prediction is a powerful post-hoc technique that takes the output of any classifier and transforms it into a set of likely labels, providing a statistical guarantee on the coverage of the true label. 
Unlike evaluations on standard benchmarks, our study addresses the complexities of  data-scarce and highly variable real-world settings. We investigate the effectiveness of leveraging pretrained models (MobileNet, DenseNet, and ResNet), fine-tuned with limited labeled data, to generate informative prediction sets. To further evaluate the impact of calibration, we consider two parallel pipelines (with and without temperature scaling) and assess performance using two key metrics: empirical coverage and average prediction set size. This setup allows us to systematically examine how calibration choices influence the trade-off between reliability and efficiency. Our findings demonstrate that even with relatively small labeled samples and simple nonconformity scores, conformal prediction can yield valuable uncertainty estimates for complex tasks. Moreover, our analysis reveals that while temperature scaling is often employed for calibration, it does not consistently lead to smaller prediction sets, underscoring the importance of careful consideration in its application. Furthermore, our results highlight the significant potential of model compression techniques within the conformal prediction pipeline for deployment in resource-constrained environments. Based on our observations, we advocate for future research to delve into the impact of noisy or ambiguous labels on conformal prediction performance and to explore effective model reduction strategies.
\end{abstract}

\section{Introduction}
\label{sec:intro}
Although deep classifiers output softmax scores for training with cross-entropy loss, their primary goal at test time is to predict the single most likely class \cite{singla2019understanding,pourkamali2023evaluation,ravi2024implicit}. Consequently, their single-label predictions lack direct uncertainty quantification. Yet, in many real-world applications, understanding a model's confidence is as important as the prediction itself \cite{corbiere2019addressing,ABDAR2021243}. Conformal prediction offers a simple yet effective post-hoc solution to generate \textit{prediction sets} from any trained classifier using calibration data \cite{shafer08a,angelopoulos2023conformal,barber2023conformal,10497110,mossina2024conformal}. These prediction sets\textemdash collections of labels guaranteed to contain the true label with a user-specified coverage level (e.g., 90\%)\textemdash provide critical information about model confidence, as smaller sets indicate higher certainty, assuming the coverage guarantee is maintained. Thus, conformal prediction transforms point predictions into confidence-informed decisions, facilitating safer deployment of machine learning models in high-stakes scenarios \cite{zhou2024conformal,singh2024uncertainty,pourkamali2024adaptive,astigarraga2025conformal}.

A key concept in conformal prediction is defining the nonconformity score function, which quantifies how ``atypical'' a given relationship between an input and its corresponding output is \cite{zargarbashi2023conformal,yuksekgonul2023beyond,kaur2025conformal,stanton2023bayesian}. Lower scores suggest higher confidence (typical examples), while higher scores indicate unusual examples. Thus, to apply conformal prediction, a calibration step follows training, where nonconformity scores are computed on a separate held-out calibration dataset. For a user-specified coverage level of $(1-\alpha)$, where $\alpha$ is a chosen level of tolerance to error (e.g., $\alpha=0.1$), a threshold is determined by calculating the $(1-\alpha)$-quantile of these calibration scores. At test time, for a new input, a prediction set is formed by considering labels that are typical for the input based on the nonconformity measure and including all classes whose nonconformity scores are below or equal to this threshold. 
 This calibration process relies on the assumption of exchangeability \cite{oliveira2024split,barber2025unifying}, meaning the calibration and test examples are assumed to be drawn from the same underlying distribution and are permutable without changing their joint distribution.

Therefore, the effectiveness of conformal prediction heavily depends on the choice of the nonconformity score function, which captures crucial information about the underlying model and the data. If the score function does not meaningfully measure atypicality or uncertainty, the resulting prediction sets will be uninformative. For instance, if the scores are random noise unrelated to the model's confidence, the prediction sets will simply contain random subsets of possible labels, large enough to achieve the desired marginal coverage but offering no useful insight. Thus, selecting a representative and meaningful score function is essential.

Since deep classifiers naturally convert the raw output scores (logits) into a probability distribution over the classes (softmax scores), a common choice of the nonconformity score function for an input-output pair is one minus the softmax score assigned to the true class (known as the Least Ambiguous set-valued Classifier or LAC \cite{sadinle2019least}). When the model assigns a softmax score of exactly one to the true class, the nonconformity score becomes zero, indicating perfect conformity and high confidence. Conversely, if the softmax probability assigned to the true class is near zero, the nonconformity score becomes large, signaling that this input-output pair is atypical and does not conform to the learned patterns of the underlying model and data. 

To achieve a desirable balance between coverage guarantees and the practical usability of prediction sets, algorithmic efforts have been directed towards designing improved score functions. For example, alternative approaches, including Adaptive Prediction Sets (APS) \cite{romano2020classification} and Regularized Adaptive Prediction Sets (RAPS) \cite{angelopoulos2020uncertainty}, consider cumulative sorted softmax probabilities rather than focusing exclusively on the probability of the true class. APS constructs prediction sets by accumulating the most probable classes until a cumulative probability threshold is reached. This adaptive method results in dynamically sized prediction sets that reflect the model's overall confidence across multiple classes. RAPS further enhances APS by introducing a regularization penalty that discourages excessively large prediction sets. 

Beyond algorithmic improvements, the urgent need for reliable prediction sets in high-stakes applications highlights the importance of thorough and realistic empirical evaluations of conformal prediction methods. Although numerous studies have empirically assessed these methods, most evaluations rely heavily on standard benchmark datasets such as MNIST, Fashion-MNIST, CIFAR-10, and CIFAR-100 (e.g., see \cite{dabah2024temperature,xi2024does,zeng2025parametric}). These datasets are large-scale and extensively preprocessed, and thus fail to fully capture the complexity and variability encountered in practical scenarios. Real-world image classification problems typically involve fewer data points\textemdash often far fewer than the tens of thousands available in standard benchmarks\textemdash and include a broader, more diverse range of classes, environmental conditions, and data-quality issues \cite{sun2021research,khoee2024domain}. Consequently, there is a need for new empirical studies utilizing datasets that better represent real-world conditions.

This work makes a significant contribution through a two-pronged approach. Empirically, we leverage the recently introduced ERA (Event Recognition in Aerial Videos) dataset \cite{mou2020era}, a meticulously human-annotated dataset of 2,864 unconstrained aerial videos made available by the remote sensing community. ERA features labels for 25 distinct dynamic event classes, each spanning a 5-second duration, and was designed with substantial intra-class variation and inter-class similarity to reflect the complexity of real-world aerial video data captured across diverse environments and scales. Recognizing the real-world difficulties of acquiring labeled aerial data in diverse and unconstrained settings (e.g., as acquired by drones), this dataset provides an excellent testbed for evaluating conformal prediction methods.

Building upon the availability of the ERA dataset, the current work undertakes a rigorous evaluation and comparison of several nonconformity score functions. To maintain comparability with existing empirical studies, our systematic approach focuses on single-frame classification, using the middle frame of each video as input image. Addressing the dataset's small scale, we employ three pretrained vision models from PyTorch\textemdash MobileNet, DenseNet, and ResNet\textemdash adapting their final layers to the ERA dataset \cite{liu2025comprehensive}. This allows us to directly address a crucial question: Can conformal prediction be applied to challenging, data-scarce problems by leveraging the transferable knowledge from pretrained vision models to produce reliable prediction sets?

\begin{figure}[htbp!]
    \centering
\includegraphics[width=\linewidth]{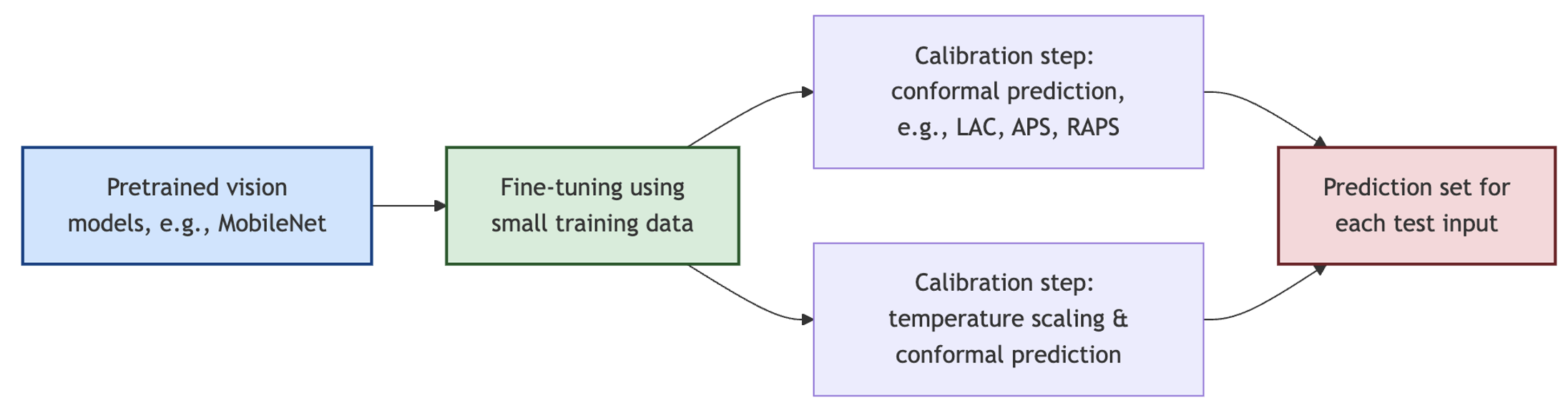}
    \caption{Illustrating the overall pipeline used in this work to evaluate the effectiveness of calibration methods, including conformal prediction and temperature scaling, on the quality of generated prediction sets in scarce and unconstrained environments.
}
    \label{fig:pipeline}
\end{figure}

Our second contribution is a systematic evaluation of the role of temperature scaling in conformal prediction. Temperature scaling is a post-hoc calibration method that adjusts the sharpness of the softmax distribution using a single parameter, tuned on a held-out calibration set \cite{guo2017calibration,sun2024logit,balanya2024adaptive}. While it is commonly used to improve the quality of uncertainty estimates, it also introduces additional complexity. We perform a comprehensive analysis of how temperature scaling affects the trade-offs between calibration accuracy and  prediction set size. Figure \ref{fig:pipeline} depicts the methodology used to generate and assess prediction sets using conformal prediction techniques.

The remainder of this paper is organized as follows. Section~\ref{sec:back} introduces the necessary notation and provides a concise overview of conformal prediction, with a focus on three widely used nonconformity score functions and incorporating temperature scaling. Section \ref{sec:benchmark} describes the experimental testbed, including the ERA dataset and the use of pretrained vision models, along with the process used to adapt these models to the ERA classification task. Section \ref{sec:exp} presents a comprehensive empirical evaluation, featuring experiments with the MobileNet architecture as well as additional results using two other pretrained models—DenseNet and ResNet. Finally, Section \ref{sec:conc} concludes the paper and outlines potential directions for future research.

\section{Notation and Preliminaries}
\label{sec:back}
In this paper, we consider a multi-class classification problem with \( C \) distinct labels. The input space is denoted by \( \mathcal{X} \), where each input \( \mathbf{x} \in \mathcal{X} \) is an image represented as a tensor in \( \mathbb{R}^{H \times W \times 3} \). The label space is defined as \( \mathcal{Y} := \{1, 2, \dots, C\} \). Classification problems aim to learn a function
$f: \mathcal{X} \rightarrow \mathcal{Y}$,
that maps inputs to their most likely class labels. Moreover, we adopt the standard supervised learning setup, assuming access to distinct labeled datasets for training and testing, denoted by \( \mathcal{D}_{\text{train}} \) and \( \mathcal{D}_{\text{test}} \), respectively. Additionally, since conformal prediction requires an intermediate calibration step, we assume the availability of a separate labeled dataset \( \mathcal{D}_{\text{cal}} \), which will be discussed in detail later in this section.

A deep neural network classifier can be viewed as a composition of multiple layers\textemdash such as convolutional, pooling, and fully connected layers\textemdash that transform the input data through increasingly abstract representations. For a classification task with $C$ classes, the final output layer contains $C$ neurons.
Conceptually, the model defines a mapping from an input \( \mathbf{x} \in \mathcal{X} \) to a vector of real-valued scores \( \mathbf{z}(\mathbf{x}) \in \mathbb{R}^C \), where each coordinate \( z_c(\mathbf{x}) \) represents the logit for class \( c \). These logits are then transformed into estimated class probabilities using the softmax function:
\begin{equation}
\pi_c(\mathbf{x}) = \frac{\exp(z_c(\mathbf{x}))}{\sum_{j=1}^C \exp(z_j(\mathbf{x}))}, \quad \text{for } c = 1, \dots, C.
\end{equation}
Hence, the softmax function exponentiates each logit, making them nonnegative, and then normalizes them by dividing by the sum of the exponentiated logits across all classes.
The resulting vector \( \boldsymbol{\pi}(\mathbf{x})=[\pi_1(\mathbf{x}),\ldots,\pi_C(\mathbf{x})] \in \mathbb{R}^C \) represents the model’s estimated conditional probabilities across all \( C \) classes. 

During the training stage, the model parameters \( \boldsymbol{\theta} \) (i.e., weights and biases) are learned by minimizing the cross-entropy loss function, which quantifies the discrepancy between the predicted probability distribution \( \boldsymbol{\pi}(\mathbf{x}) \) and the true class label \cite{pourkamali2021neural}. For each training example \( (\mathbf{x}, y) \in \mathcal{D}_{\text{train}} \), where \( y\in\{1,\ldots,C\} \) is the ground-truth label, the loss function is defined as:
\begin{equation}
\ell(\boldsymbol{\theta}; \mathbf{x}, y) = -\log \pi_{y}(\mathbf{x}),\label{eq:loss}
\end{equation}
which penalizes the model for assigning low probability to the correct class $y$. The objective during training is to find the parameters \( \boldsymbol{\theta} \) that minimize the average cross-entropy loss over all training examples:
\begin{equation}
\arg\min_{\boldsymbol{\theta}} \ \mathbb{E}_{(\mathbf{x}, y) \sim \mathcal{D}_{\text{train}}} \left[ -\log \pi_{y}(\mathbf{x}) \right].
\end{equation}
By minimizing this loss, the model learns to produce probability distributions that place higher weight on the correct labels, thereby improving its predictive accuracy.

During the testing stage, given a new input \( \mathbf{x} \) with an unknown output, the predicted class is the one with the highest softmax probability:
\begin{equation}
f(\mathbf{x}) = \arg\max_{c} \pi_c(\mathbf{x}).
\end{equation}

Although this approach produces a single predicted label, it does not provide a meaningful estimate of uncertainty \cite{minderer2021revisiting,soloff2024building}. The softmax probabilities are not guaranteed to be well-calibrated, meaning they may not reflect the true likelihood of each class. Moreover, in cases where two or more classes have similar scores, the model may select one class arbitrarily based on a marginal difference in softmax values, making the prediction brittle and potentially unstable. This lack of reliable confidence estimates can be problematic in high-stakes applications.

Conformal prediction addresses the limitations of point predictions by constructing a prediction set \( \mathcal{C}(\mathbf{x})\subseteq \mathcal{Y} \), rather than returning a single label \( f(\mathbf{x}) \). This set is expected to contain the (unknown) true label with a user-specified  error rate $\alpha$ (e.g., $\alpha=0.1$). Given a calibration dataset \( \mathcal{D}_{\text{cal}} = \{(\mathbf{x}_i, y_i)\}_{i=1}^n \) consisting of \( n \) labeled examples, the first step of conformal prediction is to define a nonconformity score function \( s(\mathbf{x}_i, y_i) \), which quantifies how atypical or incompatible the label \( y_i \) is for the input \( \mathbf{x}_i \) according to the trained classifier. A higher score indicates lower conformity, suggesting the label is less likely to be suited for the given input.

Next, we compute the nonconformity scores for all calibration examples and determine a threshold \( \hat{q} \), defined as the empirical \( \lceil (n+1)(1 - \alpha) \rceil / n \)-quantile of the scores \( \{s(\mathbf{x}_i, y_i)\}_{i=1}^n \) \cite{tibshirani2023conformal}. For a new test input \( \mathbf{x} \), the conformal prediction set \(\mathcal{C}(\mathbf{x})\) is constructed by including all class labels whose nonconformity scores do not exceed this threshold:
\begin{equation}
\mathcal{C}(\mathbf{x}) = \{ y \in \mathcal{Y} : s(\mathbf{x}, y) \leq \hat{q} \}.
\end{equation}

Intuitively, this means we include all labels that appear sufficiently ``typical'' or plausible for the input \( \mathbf{x} \) based on the calibration data. Importantly, the size of the prediction set provides a natural measure of model uncertainty: small sets (ideally singletons) suggest confident predictions, while larger sets reflect uncertainty or ambiguity in the model's output. This framework thus accommodates the fact that most classifiers are imperfect\textemdash allowing for multiple labels in difficult cases without sacrificing the desired coverage guarantee.

We now describe three widely used nonconformity score functions. Note that these scores are computed only on the calibration data, and no modifications to the training procedure are required. For a given input-label pair \( (\mathbf{x}, y) \in \mathcal{D}_{\text{cal}} \), we compute the following scores. 

\begin{itemize}
\item \textbf{Least Ambiguous Classifier (LAC)}: This nonconformity score is defined as:
    \begin{equation}
    s(\mathbf{x}, y) = 1 - \pi_y(\mathbf{x}),
    \end{equation}
    where \( \pi_y(\mathbf{x}) \) is the softmax probability assigned to the true class label \( y \). A higher softmax score\textemdash indicating greater model confidence\textemdash results in a lower nonconformity score, suggesting that the calibration pair \( (\mathbf{x}, y) \) is typical. The corresponding conformal prediction set can be written in closed form as:
    \begin{equation}
    \mathcal{C}(\mathbf{x}) = \{ y \in \mathcal{Y} : \pi_y(\mathbf{x}) \geq 1 - \hat{q} \}.
    \end{equation}
    That is, during the testing stage, we include all labels whose softmax probabilities exceed the threshold \( 1 - \hat{q} \). While LAC often yields small prediction sets, it may suffer from poor conditional coverage, particularly when the underlying classifier is miscalibrated. Additionally, a key limitation is that it can occasionally produce an empty prediction set\textemdash when none of the class probabilities surpass the required threshold.

    \item \textbf{Adaptive Prediction Sets (APS)}: APS is a nonconformity score designed to adapt the size of the prediction set based on the model's uncertainty. It considers the cumulative probability mass required to include the true label. To be precise, given the softmax probability vector \( \boldsymbol{\pi}(\mathbf{x}) = [\pi_1(\mathbf{x}), \ldots, \pi_C(\mathbf{x})] \), let \( \pi_{(1)}(\mathbf{x}) \geq \pi_{(2)}(\mathbf{x}) \geq \cdots \geq \pi_{(C)}(\mathbf{x}) \) be the sorted class probabilities in descending order. Let \( L_y \) denote the rank of the true class \( y \) in this sorted list (i.e., \( L_y = 1 \) if \( y \) has the highest probability, \( L_y = 2 \) if second highest, and so on). The APS score is then defined as:
\begin{equation}
s(\mathbf{x}, y) = \sum_{i=1}^{L_y} \pi_{(i)}(\mathbf{x}).
\end{equation}

Intuitively, this score measures how much total probability mass must be accumulated to reach the true class. A lower score indicates that the correct label appears early in the ranking (i.e., the model is more confident), while a higher score suggests greater uncertainty. 

Unlike LAC, which only considers the probability of the true class, APS accounts for the relative position of the true class among all alternatives, providing a more nuanced uncertainty measure. Prediction sets using APS are formed by including all classes until the cumulative probability reaches the threshold, making this approach naturally adaptive to the model's confidence.

\textbf{Regularized Adaptive Prediction Sets (RAPS)}: RAPS extends the APS score by incorporating an explicit regularization term that penalizes labels ranked low in the softmax ordering. The nonconformity score is defined as:
\begin{equation}
s(\mathbf{x}, y) = \sum_{i=1}^{L_y} \pi_{(i)}(\mathbf{x}) + \lambda (L_y - k_{\text{reg}})_+,
\end{equation}
where \( (\cdot)_+ = \max\{0, \cdot\} \), \( \lambda \geq 0 \) controls the strength of regularization, and \( k_{\text{reg}} \in \{1, \dots, C\} \) is a cutoff rank that determines which classes incur a penalty. 

The first term, as in APS, accumulates the softmax probabilities up to the rank of the true class, capturing the model's uncertainty. The second term introduces a penalty for true labels that appear deeper in the ranking (i.e., when \( L_y > k_{\text{reg}} \)), thereby discouraging inclusion of low-confidence labels in the prediction set. This encourages more selective prediction sets and effectively regularizes against overly uncertain predictions. RAPS is particularly useful in settings where APS may produce large prediction sets for ambiguous inputs. By tuning \( \lambda \) and \( k_{\text{reg}} \), we can control the trade-off between set size and coverage. 
\end{itemize}

A common characteristic of the nonconformity scores discussed above is their reliance on softmax probabilities, which are often miscalibrated in modern neural networks. Such miscalibration can result in unreliable uncertainty estimates and suboptimal prediction sets. To mitigate this issue, temperature scaling is frequently used as a post-hoc calibration technique. It introduces a single scalar parameter \( T > 0 \) that adjusts the confidence of predictions by rescaling the logits before applying the softmax function:
\begin{equation}
\pi_c^{(T)}(\mathbf{x}) = \frac{\exp(z_c(\mathbf{x})/T)}{\sum_{j=1}^C \exp(z_j(\mathbf{x})/T)}, \quad \text{for } c = 1, \dots, C.\label{eq:TS}
\end{equation}
The temperature \( T \) is typically tuned on a held-out calibration set by minimizing a loss function such as cross-entropy in Equation \eqref{eq:loss}, with the model parameters \( \boldsymbol{\theta} \) kept fixed after training. When \( T > 1 \), the predicted probabilities are softened to alleviate overconfidence. On the other hand, \( T < 1 \) causes the softmax output to become more ``peaked,'' concentrating higher probabilities on the the highest-logit class while reducing the probabilities assigned to other classes.

While temperature scaling is a simple and widely adopted calibration technique\textemdash often sharing the same calibration dataset \( \mathcal{D}_{\text{cal}} \) used in conformal prediction\textemdash it introduces an additional step that increases the complexity of the overall conformal prediction pipeline. In this work, we investigate the impact of temperature scaling on these nonconformity scores, including LAC, APS, and RAPS, to better understand the trade-offs between prediction set size, empirical coverage, and calibration quality (see Figure \ref{fig:pipeline}).

\section{Benchmark Environment and Model Configuration}\label{sec:benchmark}
The ERA dataset (Event Recognition in Aerial Videos) \cite{mou2020era} is a human-annotated benchmark specifically developed for recognizing events from Unmanned Aerial Vehicle (UAV) footage. It consists of 2,864 five-second video clips collected from YouTube, covering 25 event classes that span natural disasters, human activities, and routine environmental interactions. A key attribute of the ERA dataset is its focus on unconstrained, open-world settings, where videos capture diverse and often challenging scenes\textemdash ranging from urban congestion to natural terrain\textemdash at dramatically different scales and under varying environmental conditions. This makes ERA particularly suitable for developing models intended for real-world deployment, where visual variability and noise are the norm rather than the exception. In this study, we focus exclusively on the single-frame classification task, using one frame per video, which serves as a more lightweight and interpretable baseline for uncertainty-aware classification methods such as conformal prediction.

Our experiments are limited to a subset of seven event categories: Fire, Flood, Landslide, Post-Earthquake, Traffic Collision, ßConstructing, and Ploughing. This selection spans critical application areas across disaster response \cite{hajibabaee2023dimensionality}, urban planning, and precision agriculture \cite{wuepper2024satellite}. Identifying natural hazards such as fire, flood, and landslide with UAVs allows for crucial and rapid emergency assessment, especially in remote or under-monitored areas \cite{bashir2024efficient}. Traffic collisions and construction activities are central to urban analytics, offering insight into city-scale disruptions and development. Lastly, ploughing represents the agricultural domain, where UAV monitoring can enhance food security and land management. By focusing on this diverse but coherent subset of seven categories, we aim to evaluate the efficacy of
the discussed calibration methods to recognize and distinguish high-stakes events using only minimal visual information.

\begin{figure}[htbp!]
    \centering
    \includegraphics[width=0.95\linewidth]{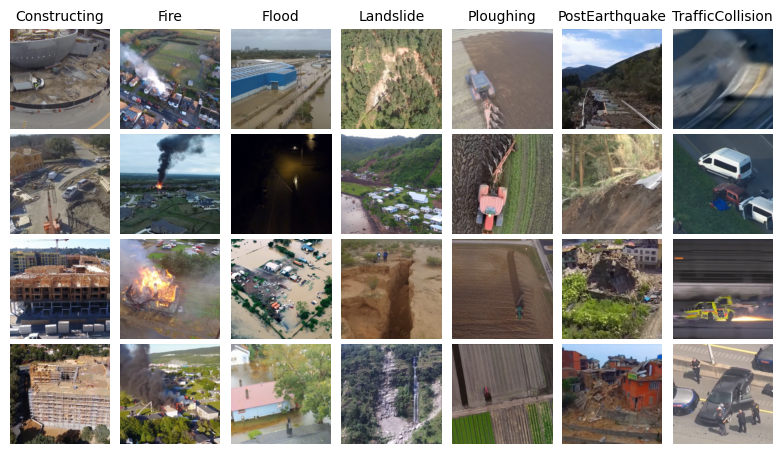}
    \caption{Four representative images from each of the seven categories used in this study are shown to illustrate the visual diversity within the dataset.}
    \label{fig:examples}
\end{figure}

The number of samples used in our study is $386$ for training, $261$ for calibration, and $112$ for testing, which is significantly fewer than the sample sizes typically used in prior conformal prediction studies. To mitigate the potential impact of any particular data split, we consider multiple random splits, which will be described in the next section.
Each image is resized to have a height of $H=224$ and a width of $W=224$ to ensure compatibility with the pretrained vision models discussed later in this section. Figure~\ref{fig:examples} displays four representative images from each selected category, highlighting the substantial visual variability present in this dataset. Consequently, this work offers a comprehensive evaluation of calibration methods in environments that are both scarce and unconstrained.

In the second part of this section, we describe the selection of pretrained vision models used as deep classifiers. Due to the limited amount of labeled data in this benchmark dataset, training large neural networks from scratch is impractical and prone to overfitting. To address this, we leverage transfer learning by utilizing three pretrained vision models available through the PyTorch library. These models offer a strong starting point by incorporating feature representations learned from large-scale image datasets, thereby improving generalization in low-data regimes.

The first model employed is MobileNetV2 \cite{sandler2018mobilenetv2}, a lightweight convolutional neural network architecture optimized for mobile and embedded vision applications. It introduces an inverted residual structure where the residual connections link thin bottleneck layers. Within each residual block, an intermediate expansion layer utilizes lightweight depthwise convolutions to filter features as a source of nonlinearity. This design reduces computational complexity while maintaining representational power, making MobileNet suitable for scenarios with limited computational resources.

The second model is DenseNet-121 \cite{arulananth2024classification}, which features a densely connected convolutional network architecture. In this design, each layer receives inputs from all preceding layers and passes its own feature maps to all subsequent layers. This dense connectivity ensures maximum information flow between layers, alleviates the vanishing-gradient problem, and strengthens feature propagation.

The third model utilized is ResNet-152 \cite{wu2019wider}, a deep residual network that addresses the degradation problem in deep neural networks. ResNet reformulates the layers as learning residual functions with reference to the layer inputs, rather than learning unreferenced functions. This is achieved through residual connections, or skip connections, that allow the network to learn identity mappings more easily, facilitating the training of much deeper networks. 

As a simple measure of model complexity, we report the total number of trainable parameters in each pretrained architecture: MobileNetV2 has approximately 3.5 million parameters, DenseNet-121 has 7.98 million, and ResNet-152 contains 60.19 million. Given that the benchmark dataset includes \( C = 7 \) event categories, we replace the final classification layer of each model with a fully connected layer consisting of seven output neurons. To limit overfitting, all other layers are frozen, and only the newly added classification head is trained. Each model is fine-tuned for 10 epochs using a batch size of 8 and a learning rate of 0.001, with the cross-entropy loss function. These classifiers remain fixed during the subsequent calibration stage.

\section{Conformal Prediction Performance}\label{sec:exp}
In this section, we use the ERA dataset to investigate two critical components of the calibration stage for generating prediction sets: (1) the choice of the nonconformity score function, and (2) the effect of temperature scaling. To this end, we evaluate the three methods introduced in Section~\ref{sec:back}, namely LAC, APS, and RAPS. As a brief reminder, LAC computes nonconformity scores using only the softmax probability of the true class, whereas APS and RAPS first sort the softmax scores and then incorporate the rank of the true label within this ordering.

We consider two calibration scenarios: without and with temperature scaling (TS). In the baseline case (without TS), softmax scores are computed directly from the model logits to form a $C$-dimensional probability vector $\boldsymbol{\pi}(\mathbf{x})$. When applying TS, we introduce a single scalar parameter \( T \) during calibration, which rescales the logits before applying the softmax function to compute $\boldsymbol{\pi}^{(T)}(\mathbf{x})$; see Equation \eqref{eq:TS}. The value of \( T \) is optimized by minimizing the cross-entropy loss on the calibration set \( \mathcal{D}_{\text{cal}} \), while keeping all network parameters fixed. This approach aims to improve the calibration of softmax outputs prior to constructing conformal prediction sets. Hence, we provide a comprehensive analysis of how temperature scaling (denoted by TS) influences the performance of different nonconformity scoring methods in conformal prediction.

For implementation, we rely on the PyTorch deep learning framework to access pretrained vision models and perform fine-tuning using limited labeled data. PyTorch provides a flexible and modular environment for loading pretrained weights and modifying network architectures, which is essential for adapting the final classification layer to our specific task. To integrate with libraries designed for conformal prediction, we wrap our PyTorch models using a scikit-learn-compatible interface via the skorch library \cite{skorch}. This compatibility is crucial for leveraging the MAPIE (Model Agnostic Prediction Interval Estimator) library \cite{Cordier_Flexible_and_Systematic_2023}, which requires estimators to conform to the scikit-learn API. The MAPIE library enables efficient implementation of various conformal prediction methods\textemdash including LAC, APS, and RAPS\textemdash by automating score computation, quantile calibration, and prediction set construction in a model-agnostic manner.

In each experiment, we evaluate the quality of the generated prediction sets using two key metrics: empirical coverage and average prediction set size. Recall that we specify an error rate parameter \( \alpha \) (in this paper, we use \( \alpha = 0.2 \) and \( \alpha = 0.1 \)), which corresponds to a theoretical coverage level of \( 1 - \alpha \) (thus, $0.8$ and $0.9$, respectively). To assess whether this desired coverage is achieved in practice, we compute the \textit{empirical coverage}, defined as the proportion of test examples in \( \mathcal{D}_{\text{test}} \) for which the prediction set contains the true label.

The second metric is the average prediction set size, which reflects the model's confidence in its predictions. In general, smaller prediction sets are preferable, as they indicate higher certainty, whereas larger sets suggest greater uncertainty. Since our task involves \( C = 7 \) possible labels, a trivial method could generate uninformative prediction sets by simply including approximately \( (1 - \alpha) \times C \) labels at random. For instance, when \( \alpha = 0.2 \), this corresponds to a prediction set size of 5.6 on average\textemdash achieving the nominal coverage but offering little discriminatory power. Therefore, effective conformal methods should produce prediction sets that are not only valid (i.e., achieving the desired coverage) but also much smaller than this trivial baseline. 

In addition to these two metrics, we also report supplementary measures to provide a comprehensive analysis of the calibration stage, including the distribution of the optimized temperature parameters $T$ and visualizations of selected test images alongside their true labels and corresponding prediction sets.

In the first round of our experiments, we focus on MobileNet due to its lightweight architecture and relatively low number of trainable parameters, making it computationally efficient and well-suited for scenarios with limited resources. For each experimental condition, we repeat the entire training, calibration, and testing pipeline, depicted in Figure \ref{fig:pipeline}, across 50 independent trials and report the variability observed in the results. 

This repeated evaluation serves two important purposes. First, neural network training is inherently stochastic due to factors such as random weight initialization, data shuffling, and mini-batch selection \cite{gundersen2022sources,semmelrock2025reproducibility}. By averaging over multiple trials, we mitigate the impact of this randomness and obtain more reliable performance estimates. Second, in each trial, we use a different random split of the data into training, calibration, and testing subsets; see the pipeline in Figure \ref{fig:pipeline}. This helps assess the robustness of each method to variability in the dataset partitioning and ensures that the reported performance is not overly dependent on a particular data split. Together, these repeated trials provide a comprehensive and statistically sound evaluation of both the deep classifier and the calibration methods under consideration.

\begin{figure}[htbp!]
    \centering
\includegraphics[width=\linewidth]{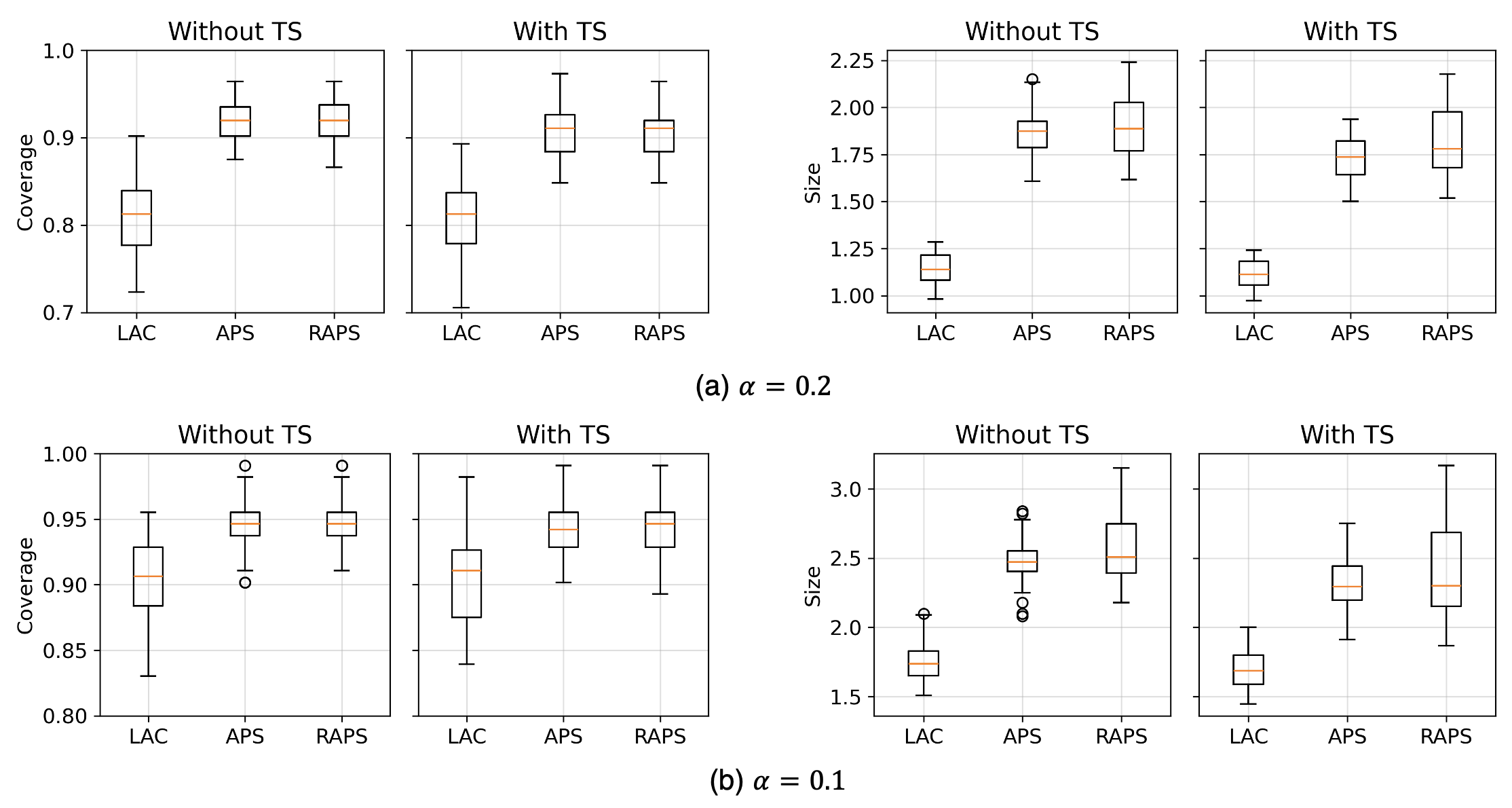}
    \caption{Boxplots of coverage scores and prediction set sizes across 50 independent trials using the MobileNet architecture, shown for two error rate values: (a) $\alpha = 0.2$ and (b) $\alpha = 0.1$. TS denotes temperature scaling, which is applied prior to computing the nonconformity score functions to evaluate its impact on calibration performance.}
    \label{fig:mobilenet}
\end{figure}

Figure \ref{fig:mobilenet} presents boxplots of the empirical coverage scores and average prediction set sizes for the three conformal prediction methods, evaluated at two error rates: $\alpha = 0.2$ and $\alpha = 0.1$. As shown in Figure \ref{fig:mobilenet}(a), the median empirical coverage for all three methods exceeds the target level of $1-0.2=0.8$. However, the first quartile of LAC's coverage scores falls below this threshold, indicating that LAC may yield lower-than-expected coverage in some cases. In contrast, even the minimum coverage scores for APS and RAPS are well above the target level based on the left panel of Figure \ref{fig:mobilenet}(a). For instance, across 50 trials without applying temperature scaling (TS), the minimum coverage for both APS and RAPS is approximately 0.87. This value drops slightly to around 0.85 when TS is applied prior to conformal prediction. Despite this minor reduction, the coverage performance of all three methods remains comparable with and without TS when $\alpha = 0.2$.

Next, we examine the average prediction set sizes across 50 trials when $\alpha = 0.2$, as shown in the right panel of Figure \ref{fig:mobilenet}(a). As expected, LAC yields the smallest prediction sets, with median sizes of 1.14 without TS and 1.11 with TS. These results demonstrate that LAC is capable of producing informative and compact prediction sets while still maintaining the desired coverage of $0.8$. In contrast, the higher coverage scores achieved by APS and RAPS come at the cost of noticeably larger prediction sets. For instance, with TS applied, the median prediction set sizes for APS and RAPS are 1.74 and 1.78, respectively. Notably, applying TS leads to a reduction in prediction set sizes across all methods in this experiment, which is a desirable outcome and highlights the practical benefit of including the temperature scaling step in the pipeline.

After discussing the results for $\alpha = 0.2$, we now turn to the results presented in Figure \ref{fig:mobilenet}(b) for $\alpha = 0.1$. Once again, we observe that the median coverage scores for all three conformal prediction methods reach the target level of $0.9$, regardless of whether temperature scaling is applied. However, similar to the previous case, LAC exhibits more variability and can produce less reliable coverage, with minimum scores of 0.83 and 0.84 without and with TS, respectively. In contrast, APS and RAPS demonstrate more robust performance, with the majority of coverage scores well above 0.9 and median values slightly below 0.95. Finally, we note that incorporating TS as an additional calibration step does not substantially affect the coverage performance.

Next, we compare the prediction set sizes when $\alpha = 0.1$, as shown in the right panel of Figure \ref{fig:mobilenet}(b). The prediction set sizes for LAC increase compared to those observed at $\alpha = 0.2$. For example, with temperature scaling (TS), the median prediction set sizes for LAC are 1.11 and 1.69 at $\alpha = 0.2$ and $\alpha = 0.1$, respectively. This increase is anticipated, as a higher desired coverage level naturally leads to larger prediction sets. Nevertheless, the prediction sets produced by LAC remain highly informative and compact; even at $\alpha = 0.1$, the maximum prediction set size remains close to 2, indicating that, on average, only 2 out of 7 possible categories are sufficient to achieve 90\% coverage. In contrast, APS and RAPS result in larger prediction sets than LAC when $\alpha=0.1$, reflecting their higher empirical coverage levels. Interestingly, applying TS slightly reduces the median prediction set sizes for both APS and RAPS, with values close to 2.3. 

These observations underscore the trade-off between coverage and prediction set size. In high-stakes applications\textemdash where achieving or exceeding the desired coverage is critical\textemdash APS and RAPS offer a dependable solution: when applied to a fine-tuned MobileNet model, they consistently meet the target coverage while maintaining relatively compact prediction sets, typically containing only 2 to 3 labels. However, if the primary goal is to minimize prediction set size, LAC offers a more concise and potentially more informative alternative.

In the next set of experiments, we focus on comparing the performance of conformal prediction methods across the three deep classifiers discussed in the previous section: MobileNet, DenseNet, and ResNet. First, we report the coverage scores in Table~\ref{table:cov} for two values of $\alpha$: 0.2 and 0.1. Consistent with the earlier boxplots, we observe that the mean coverage scores for LAC closely approach the target level of $1 - \alpha$. In contrast, APS and RAPS tend to exceed the desired coverage, yielding empirical values around $0.90$ for $\alpha = 0.2$ and approximately $0.95$ for $\alpha = 0.1$. These results indicate that APS and RAPS surpass the target coverage across all three models. 

\begin{table}[ht]
\centering
\caption{Mean $\pm$ standard deviation of coverage scores across 50 trials for different models, conformal prediction methods, and error rate values $\alpha$, both with and without temperature scaling (TS).\label{table:cov}}
\resizebox{\textwidth}{!}{%
\begin{tabular}{llcccccc}
\toprule
\multirow{2}{*}{Model} & \multirow{2}{*}{$\alpha$} & \multicolumn{2}{c}{LAC} & \multicolumn{2}{c}{APS} & \multicolumn{2}{c}{RAPS} \\
\cmidrule(lr){3-4} \cmidrule(lr){5-6} \cmidrule(lr){7-8}
 & & Without TS & With TS & Without TS & With TS & Without TS & With TS \\
\midrule
\multirow{2}{*}{MobileNet} & 0.2 & 0.81$\pm$0.05 & 0.81$\pm$0.05 & 0.92$\pm$0.02 & 0.91$\pm$0.03 & 0.92$\pm$0.02 & 0.91$\pm$0.03 \\
                           & 0.1 & 0.91$\pm$0.03 & 0.90$\pm$0.03 & 0.95$\pm$0.02 & 0.94$\pm$0.02 & 0.95$\pm$0.02 & 0.94$\pm$0.02 \\
\addlinespace
\multirow{2}{*}{DenseNet} & 0.2 & 0.81$\pm$0.04 & 0.81$\pm$0.04 & 0.90$\pm$0.02 & 0.90$\pm$0.02 & 0.90$\pm$0.02 & 0.90$\pm$0.03 \\
                          & 0.1 & 0.90$\pm$0.03 & 0.90$\pm$0.03 & 0.94$\pm$0.02 & 0.94$\pm$0.02 & 0.94$\pm$0.02 & 0.94$\pm$0.02 \\
\addlinespace
\multirow{2}{*}{ResNet}   & 0.2 & 0.80$\pm0.04$ & 0.80$\pm0.04$ & 0.89$\pm0.03$ & 0.91$\pm0.03$ & 0.89$\pm0.03$ & 0.90$\pm0.03$\\
                          & 0.1 & 0.89$\pm0.03$ & 0.89$\pm0.03$ & 0.93$\pm0.03$ & 0.94$\pm0.02$ & 0.93$\pm0.03$ & 0.94$\pm0.02$ \\
\bottomrule
\end{tabular}
}
\end{table}

Regarding the impact of temperature scaling, we observe that its effect on coverage scores is minimal, with differences typically within 1\% compared to the results obtained without TS. Finally, it is worth noting that the three classifiers exhibit comparable performance across the conformal prediction methods. For example, despite substantial differences in model size and complexity, such as between MobileNet and ResNet, there is no striking variation in coverage scores, highlighting the robustness of these methods across architectures.

As noted earlier, both coverage scores and average prediction set sizes must be considered together to evaluate the performance of conformal prediction methods effectively. Table~\ref{table:size} reports the mean and standard deviation of prediction set sizes. LAC consistently produces smaller prediction sets compared to APS and RAPS. For instance, using MobileNet with temperature scaling, the mean prediction set sizes for LAC are 1.11 and 1.70 for $\alpha = 0.2$ and $\alpha = 0.1$, respectively. These results demonstrate that it is possible to obtain informative and compact prediction sets even on a scarce and unconstrained benchmark dataset. Similar trends are observed with DenseNet, where the prediction set sizes slightly increase to 1.14 and 1.79 for the respective values of $\alpha$. Overall, these findings indicate that the performance of conformal prediction methods is not overly sensitive to the choice of classifier, which is a desirable property for real-world deployment.

\begin{table}[ht]
\centering
\caption{Mean $\pm$ standard deviation of prediction set sizes across 50 trials for different models, conformal prediction methods, and error rate values $\alpha$, both with and without temperature scaling.\label{table:size}}
\resizebox{\textwidth}{!}{%
\begin{tabular}{llcccccc}
\toprule
\multirow{2}{*}{Model} & \multirow{2}{*}{$\alpha$} & \multicolumn{2}{c}{LAC} & \multicolumn{2}{c}{APS} & \multicolumn{2}{c}{RAPS} \\
\cmidrule(lr){3-4} \cmidrule(lr){5-6} \cmidrule(lr){7-8}
 & & Without TS & With TS & Without TS & With TS & Without TS & With TS \\
\midrule
\multirow{2}{*}{MobileNet} & 0.2 & 1.14$\pm$0.08 & 1.11$\pm$0.07 & 1.86$\pm$0.12 & 1.74$\pm$0.11 & 1.91$\pm$0.15 & 1.82$\pm$0.18 \\
                           & 0.1 & 1.75$\pm$0.13 & 1.70$\pm$0.14 & 2.48$\pm$0.16 & 2.30$\pm$0.18 & 2.59$\pm$0.26 & 2.42$\pm$0.38\\
\addlinespace
\multirow{2}{*}{DenseNet} & 0.2 & 1.16$\pm0.08$ & 1.14$\pm0.08$ & 1.89$\pm0.13$ & 1.78$\pm0.14$ & 1.99$\pm0.17$ & 1.85$\pm0.17$ \\
                          & 0.1 & 1.84$\pm0.23$ & 1.79$\pm0.21$ & 2.61$\pm0.24$ & 2.45$\pm0.24$ & 2.82$\pm0.36$ & 2.66$\pm0.50$ \\
\addlinespace
\multirow{2}{*}{ResNet}   & 0.2 & 1.06$\pm$0.09  & 1.06$\pm$0.09 & 1.70$\pm$0.16 & 1.83$\pm$0.19 & 1.69$\pm$0.19 & 1.84$\pm$0.20 \\
                          & 0.1 & 1.62$\pm$0.26 & 1.59$\pm$0.24 & 2.29$\pm$0.28 & 2.50$\pm$0.36 & 2.45$\pm$0.39 & 2.63$\pm$0.47 \\
\bottomrule
\end{tabular}
}
\end{table}

Furthermore, we observe a somewhat different pattern for the ResNet classifier. In this instance, the impact of temperature scaling on LAC appears negligible. However, applying TS leads to larger mean prediction set sizes for both APS and RAPS. For example, with $\alpha=0.1$, the mean prediction set sizes for APS without and with TS are 2.29 and 2.50, respectively, representing approximately a 9\% increase in size. Therefore, our analysis reveals a crucial aspect of temperature scaling: it does not invariably guarantee smaller prediction sets. Consequently, it is generally advisable to consider the impact of temperature scaling during the calibration stage when employing conformal prediction.

\begin{figure}[htbp!]
    \centering
    \includegraphics[width=\linewidth]{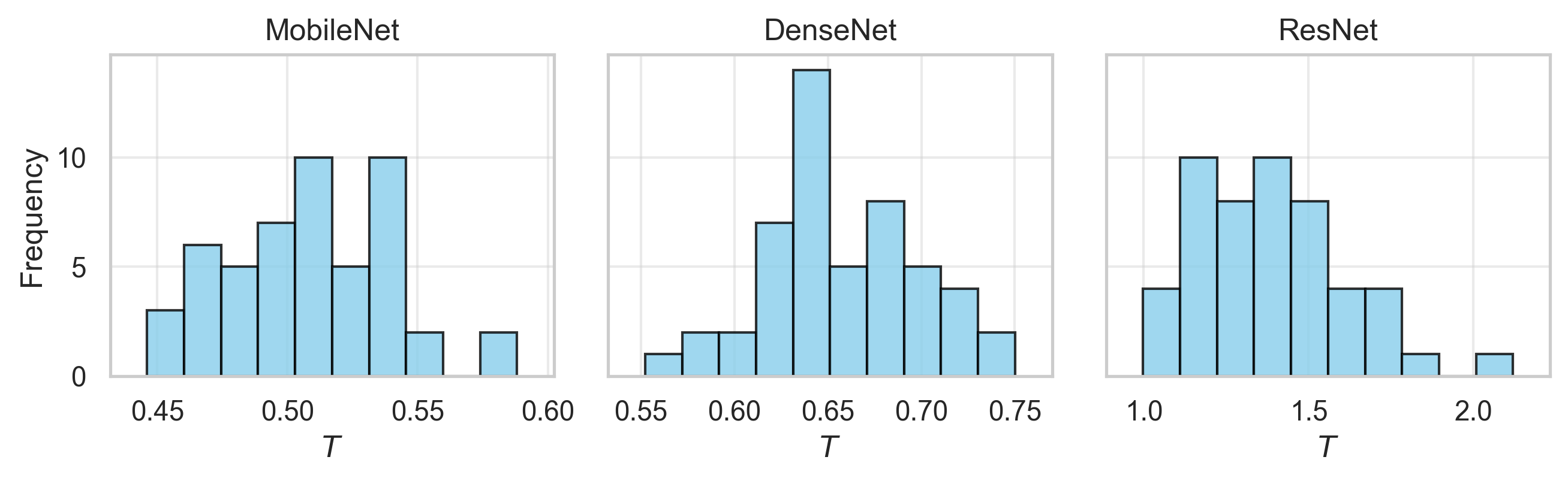}
    \caption{Histogram plots of the optimized temperature parameter values during the calibration stage for the three classifiers.}
    \label{fig:temperature_hist}
\end{figure}

To further investigate the distinction between the first two models (MobileNet and DenseNet) and the third model (ResNet), we present histograms of the optimized temperature parameters $T$ across 50 independent trials in Figure~\ref{fig:temperature_hist}. These results reveal an interesting pattern: the optimized temperature values for MobileNet and DenseNet are consistently less than 1, whereas for ResNet, they are greater than 1. This suggests that, during calibration, temperature scaling sharpened the softmax scores for MobileNet and DenseNet but softened them for ResNet. This contrast provides insight into the earlier observation that applying temperature scaling to ResNet led to increased prediction set sizes for APS and RAPS. By softening the softmax outputs, temperature scaling reduces the relative confidence of the top-ranked classes, resulting in higher nonconformity scores and thus larger prediction sets. In contrast, sharpening the outputs (as in MobileNet and DenseNet) increases confidence in the top classes, which can lead to smaller prediction sets while maintaining the coverage guarantee. Therefore, the direction of the temperature adjustment, i.e.,  sharpening versus softening, plays a key role in how temperature scaling influences the size of the prediction sets. 

\begin{figure}
    \centering
    \includegraphics[width=0.95\linewidth]{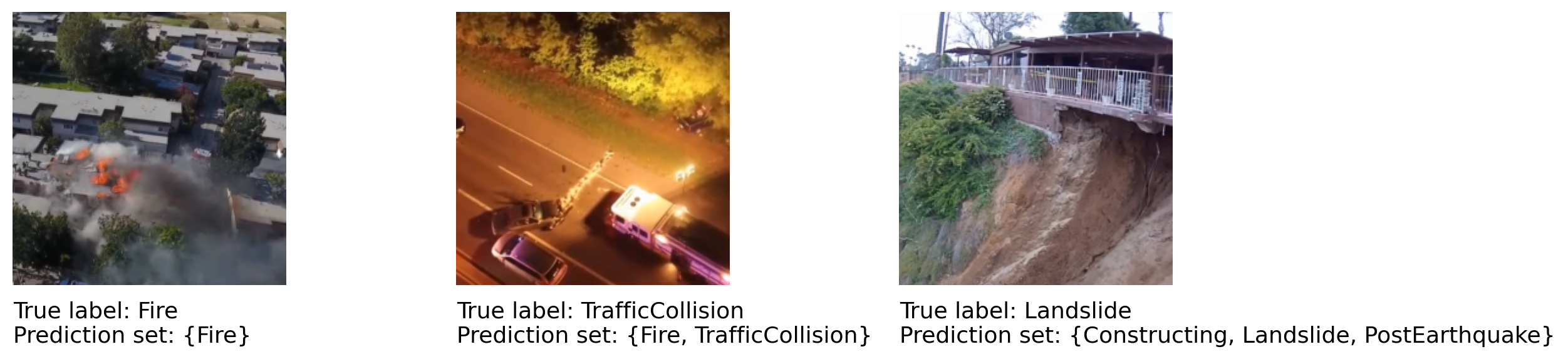}
    \caption{Three test images are shown with their true labels and the corresponding prediction sets produced by LAC.}
    \label{fig:images}
\end{figure}

Overall, this section has demonstrated that our simplest deep classifier combined with the simplest conformal prediction method\textemdash specifically, MobileNet and LAC\textemdash yield informative and compact prediction sets. To further examine these sets beyond the coverage score and average prediction set size across all test points, Figure \ref{fig:images} visualizes three test images, their true labels, and their generated prediction sets for $\alpha=0.1$ without temperature scaling. The leftmost image depicts an aerial view of a fire event, and its corresponding prediction set contains only the correct label. This is logical, as the image clearly shows a fire event with little ambiguity. The middle image distinctly shows a traffic accident; however, the lighting in this image also makes it resemble a fire event. Consequently, the prediction set obtained by LAC reflects this potential ambiguity, including both ``Fire'' and ``TrafficCollision''. 

The rightmost image presents the most intriguing case. Although labeled as ``landslide'' in the dataset, it exhibits characteristics that could also suggest ``Constructing'' or ``PostEarthquake''. Therefore, we observe that conformal prediction methods can indeed provide meaningful prediction sets that adapt to the context of each image. Prediction sets containing more than one label are particularly valuable for further examination by human and domain experts.

\section{Conclusion and Future Research}\label{sec:conc}
This paper presented a comprehensive evaluation of three widely used conformal prediction methods on a novel and challenging benchmark: an aerial image dataset of various events captured in unconstrained environments. This provided a more realistic assessment compared to evaluations on simpler, larger datasets with fewer complexities. Our investigation into the role of the classifier model demonstrated the effectiveness of fine-tuning pretrained vision models (MobileNet, DenseNet, and ResNet) with surprisingly limited labeled data (less than 400 samples) to achieve informative prediction sets.

Furthermore, our examination of temperature scaling in the calibration process revealed a nuanced relationship between the optimized temperature parameter and prediction set size. We showed that while temperature scaling can often lead to smaller sets, it can also unexpectedly increase them, particularly when the optimized temperature is above one. This underscores the necessity for a more thorough understanding of temperature scaling's impact in conformal prediction applications.

Importantly, our findings indicated that even relatively parameter-efficient models like MobileNet, coupled with a simple softmax-based nonconformity score, yielded informative and compact prediction sets for this demanding problem. This is a significant result, suggesting that complex models and sophisticated nonconformity measures are not essential for achieving valuable conformal prediction outcomes.

Building upon these observations, we recommend two primary directions for future research. First, a deeper investigation into the effects of ambiguous or noisy labels in the training or calibration sets on conformal prediction performance, requiring dedicated algorithmic and empirical efforts, such as extensions of the experiments presented in \cite{einbinder2024label}. Second, the exploration and integration of model reduction techniques, such as knowledge distillation \cite{gou2021knowledge,pourkamali2021adaptive}, within the conformal prediction calibration step to improve computational efficiency. These research directions hold significant promise for enabling the practical application of conformal prediction in high-stakes and resource-limited settings, such as aerial platforms for real-time disaster monitoring.

\bibliographystyle{plain}
\bibliography{refs}

\end{document}